# Adaptive Neural Network Ensemble Using Frequency Distribution

Ungki Lee

Cho Chun Shik Graduate School of Mobility,

Korea Advanced Institute of Science and Technology,

Daejeon, 34051, Republic of Korea

lwk920518@kaist.ac.kr

Namwoo Kang*

Cho Chun Shik Graduate School of Mobility,

Korea Advanced Institute of Science and Technology,

Daejeon, 34051, Republic of Korea

nwkang@kaist.ac.kr

* Corresponding author

**Abstract**

Neural network (NN) ensembles can reduce large prediction variance of NN and improve prediction accuracy. For highly nonlinear problems with insufficient data set, the prediction accuracy of NN models becomes unstable, resulting in a decrease in the accuracy of ensembles. Therefore, this study proposes a frequency distribution-based ensemble that identifies core prediction values, which are expected to be concentrated near the true prediction value. The frequency distribution-based ensemble classifies core prediction values supported by multiple prediction values by conducting statistical analysis with a frequency distribution, which is based on various prediction values obtained from a given prediction point. The frequency distribution-based ensemble can improve predictive performance by excluding prediction values with low accuracy and coping with the uncertainty of the most frequent value. An adaptive sampling strategy that sequentially adds samples based on the core prediction variance calculated as the variance of the core prediction values is proposed to improve the predictive performance of the frequency distribution-based ensemble efficiently. Results of various case studies show that the prediction accuracy of the frequency distribution-based ensemble is higher than that of Kriging and other existing ensemble methods. In addition, the proposed adaptive sampling strategy effectively improves the predictive performance of the frequency distribution-based ensemble compared with the previously developed space-filling and prediction variance-based strategies.

# 1. Introduction

A surrogate model is built to represent the true model with data obtained from a limited number of simulations or experiments; thus, predictions for computationally expensive models can be approximated (Kang et al., 2019). Surrogate modeling focuses on input–output behavior to find a model that approximates the relationship between input and output as accurately as possible. Neural network (NN) is one of the promising surrogate models and is known to be a universal function approximator with a good prediction accuracy (Pan et al. 2014). Multiple artificial neurons that mimic the structure and principle of biological neurons constitute NN; these artificial neurons are interconnected to form a network that derives output for given input data (Basheer and Hajmeer 2000). NN comprises layered structures, in which nodes in each layer are connected to nodes in subsequent layers and information moves in the forward direction. NN parameters include weights and biases, and weights are given to connections between nodes (Eason and Cremaschi 2014). The weighted sum of inputs from each node is transferred to the next node through an activation function, and the parameters can be determined through training. Once a given data set is split into training and validation data set, the training data set is used to train the NN model and the validation data set is used to estimate the accuracy of the trained model. Applying NN as a surrogate model in various engineering domains, such as mechanical engineering (Chen et al. 2020; Lin et al. 2020; Ktari et al. 2021; Schrader and Schauer 2021), structural engineering (Gomes et al., 2011; de Santana Gomes 2019; Yılmaz et al. 2021; Freitag et al. 2020), aerospace engineering (Bouhlel et al. 2020; Du et al. 2021; Zhang et al. 2021), biomedical engineering (Lu et al. 2013; Eskinazi and Fregly 2015), chemical engineering (Eason and Cremaschi 2014; Moreno-Pérez et al. 2018), civil engineering (Shaw et al. 2017; García-Segura et al. 2017; Thrampoulidis et al. 2021), and composite structures (Papadopoulos et al. 2018; Yan et al. 2020), has been recently attempted.

However, NN has a problem in that different models are created in accordance with the selection of training data sets, initial parameters, and training algorithms (Zhang 2007). This condition results in various prediction values; thus, NNs are referred to have high prediction variance. Ensembles that combine prediction values obtained from multiple component models have been developed to reduce the prediction variance of NNs (Sollich and Krogh 1996). The basic premise of ensembles is that the errors of a single model can be compensated by the other models (Sagi and Rokach 2018). Ensembles are generally constructed through two steps: training multiple component models and combining prediction values derived from the multiple component models. Sufficiently diverse component models with high predictive performance are required for the ensemble to have high accuracy (Deng et al. 2013). Various component models can be created depending on the combination of training and validation data sets and the selection of initial parameters. If the validation error of the model is small, then the predictive performance can be estimated to be accurate. The accuracy of the ensemble is highly dependent on the combination of prediction values. Thus, various methods, such as average ensemble (Opitz and Shavlik 1996), weighted ensemble (Bishop 1995), and mode ensemble (Kourentzes et al. 2014), which combine prediction values, have been developed. Many studies have shown that the prediction accuracy of ensembles combining multiple component models is higher than that of a single component model (Wolpert 1992; Goodfellow et al. 2016).

In the case of real engineering applications, the size of the data set for creating surrogate models is usually small because simulations or experiments (e.g., finite element analysis and collision test) are time consuming and expensive (Gaspar et al. 2017; Lee et al. 2020). For highly nonlinear problems, the prediction accuracy of NN decreases, and the situation worsens when the size of the data set is small. The deviation of prediction values increases as the prediction accuracy of component models becomes unstable, resulting in a decrease in the accuracy of ensembles. The mode ensemble that can cope well with outliers is suitable when the deviation of prediction values is large. In the mode ensemble, kernel density estimation (KDE) is performed on the prediction values, and the value corresponding to the maximum density is identified as a mode and used as the final prediction value (Kourentzes et al. 2014). The mode is determined on the basis of the most frequent value; thus, the mode ensemble is insensitive to outliers compared with average and weighted ensembles. However, a biased prediction value can be obtained in the presence of multiple high-frequency values. In addition to the most frequent value, keeping the possibility open to various prediction values is also required considering the unstable prediction accuracy.

Therefore, this study proposes a frequency distribution-based ensemble that explores the range of core prediction values, which are expected to be close to the true prediction value. For a given prediction point, various prediction values obtained from the component models constitute a prediction value distribution, and the frequency distribution-based ensemble performs statistical analysis with a frequency distribution based on the prediction value distribution to identify the core prediction values. Prediction values belonging to the range supported by multiple prediction values can be classified into core prediction values, and the average of the core prediction values is determined as the final prediction value. An adaptive sampling strategy, which efficiently improves the accuracy of the frequency distribution-based ensemble according to the variance of core prediction values, is also proposed. Various examples are used to verify the proposed frequency distribution-based ensemble and adaptive sampling strategy.

The rest of this paper is organized as follows. Section 2 reviews previous works related to NNs and ensembles.

Section 3 presents the proposed frequency distribution-based ensemble and the adaptive sampling strategy, which efficiently improves the predictive performance of the proposed frequency distribution-based ensemble. Section 4 demonstrates the predictive accuracy of ensemble methods and compares the proposed adaptive sampling strategy with other existing sampling strategies through case studies. Section 5 finally provides the conclusion and future directions.

## 2. Related works

### 2.1. Application of neural networks (NNs)

NNs inspired by the human brain mimic the behavior of biological neurons and have been successfully used in clustering, classification, pattern recognition, and regression (Abiodun et al. 2018). Five processes mainly constitute NN modeling (Öztaş et al. 2006): (1) problem definition and data collection; (2) NN architecture determination; (3) training process determination; (4) NN training; and (5) NN validation. An input layer, multiple hidden layers, and an output layer constitute the structure of the NN, and the nodes of each layer are connected to those of the next layer based on weights. The input layer is a non-computational layer that only receives input data, whereas linear or nonlinear computations are performed in the hidden and output layers, respectively (Golafshani et al. 2020). Once the inputs are transmitted between nodes, the weighted sum of inputs is passed along with the bias to the activation function and the output is computed. The NN model is trained to find optimal weights and biases that can minimize the loss based on the dataset comprising input and output data (Tan et al. 2021). The advantages of NNs are high accuracy and a fast process using parallel computing (Izeboudjen et al. 2014).

NNs can be used as surrogate models for analysis and prediction involving time-consuming simulations or expensive experiments. Eskinazi and Fregly (2015) proposed a novel surrogate contact modeling method based on NN to replace deformable joint contact models involving computationally expensive contact evaluations. The proposed method solves the computational bottleneck of musculoskeletal simulations or optimizations that require deformable joint contact models. Kumar et al. (2018) adopted NN to predict the wear of Al6061 alloy reinforced with aluminum oxide particulates. Wear height reduction according to applied load, sliding distance, and weight percentage of reinforcement is approximated through NN, and the results of experiments and the NN model predictions are in good agreement. García-Alba et al. (2019) used NN to reduce the computational cost required for the analysis of fecal indicator organism concentrations in estuaries. Maleki et al. (2019) predicted the spatial–temporal profile of pollutant concentrations and air quality indexes using NN. Liu et al. (2021) implemented NN to model chloride diffusion coefficient associated with chloride ingression, which is a major cause of corrosion of reinforced concrete structures. This study verifies that the predicted diffusion coefficient results are well matched with the experimental results.

Design optimization entails numerous function evaluations or simulations in objective function evaluation and sensitivity analysis, respectively; thus, many studies have adopted NN as a surrogate model. Shaw et al. (2017) maximized the hydropower generation of the dam using a high-fidelity hydrodynamic and water quality model approximated by NN. Bouhlel et al. (2020) proposed gradient-enhanced NNs, which improve the accuracy of the NN model with gradient information, to reduce the computational time required for analysis and design optimization of airfoil shape design. The proposed method yields similar results to optimization designs based on high-fidelity computational fluid dynamics. In heat source layout optimization problem, Chen et al. (2020) evaluated the thermal performance according to the input layout by mapping the layout and the corresponding temperature field using a feature pyramid network, which is a kind of NN. Lin et al. (2020) optimized the design of the automotive semi-active suspension system by employing radial basis function neural network, which approximates the regulating mechanism of the hydraulic adjustable damper. Yılmaz et al. (2021) used NN to the continuous motorcycle protection barrier design to reduce weight and cost. Zhu et al. (2022) approximated thermoelectric module simulation employing NN to optimize a thermoelectric generator design to improve the power generation performance.

In the field of reliability-based design optimization (RBDO), which derives a highly reliable design considering the uncertainties of engineering systems, numerous function evaluations or simulations are required for the reliability calculation; thus, NN-based surrogate models have been frequently applied (Elhewy et al. 2006). For reliability analysis of structures, Cheng and Li (2008) developed an artificial neural network-based genetic algorithm (ANN-GA), which approximates the limit state function with an ANN model and calculates the failure probability with the GA. Papadopoulos et al. (2012) proposed a reliability analysis method that combines NN and subset simulation (SS). In the proposed method, NNs are trained on multiple subdomains generated from SS and used as surrogate models to increase the efficiency of SS. Lehký et al. (2018) presented an approach to solve RBDO based on the inverse reliability method using NN. Nezhad et al. (2019) introduced a group method of data handling-type NNs with general structure for the reliability analysis of structures. Lee et al. (2022) proposed a new inverse reliability analysis method combining NN with Monte Carlo simulation. In the proposed method, NN is used to train the relationship between the small number of samples and the corresponding true percentile value. Ren et al. (2022) presented a local best surrogate (LBS) approach and a local weighted average surrogate (LWAS) approach that combine NNs and Kriging for structural reliability analysis. The LBS approach selects the surrogate model with minimum predicted error, and the LWAS approach assigns weights to the surrogate models according to the prediction error. NNs and Kriging are used in both approaches as candidate surrogate models.

**2.2. Ensembles**

Ensembles that combine prediction values derived from different NN models have been widely studied to improve the prediction accuracy and robustness of NNs, and using ensembles for prediction using NNs is reasonable (Crone et al. 2011). The predictive performance of ensembles varies depending on the combination of individual predictions (Stock and Watson 2004). Ensembles began as an attempt to improve forecasting accuracy through model combinations in various surrogate modeling methods. Bates and Granger (1969) showed that the combination of forecasts can improve forecasting accuracy. Newbold and Granger (1974) found that improvements in forecasting accuracy can be obtained by linearly combining individual forecasts in univariate time series forecasting. Dasarathy and Sheela (1978) proposed a composite classifier system comprising multiple component classifiers to improve recognition system performance. Makridakis and Winkler (1983) investigated the improvement of forecasting accuracy by averaging forecasts obtained from several individual methods and confirmed that the accuracy depends on the number of individual methods and the specific methods used for averaging. Hansen and Salamon (1990) showed that a plurality consensus scheme, which selects the one supported by more networks than anyone else, improves the predictive performance of NN ensembles. Krogh and Vedelsby (1995) proposed a weighted average of NNs based on generalization error. In weighted forecast combinations, Elliott and Timmermann (2004) found that optimal weights vary in accordance with the underlying forecast error distribution. The details of previously developed ensemble methods for NNs will be covered in Section 4.1.1.

Ensembles have been verified to improve predictive performance; thus, NN ensembles have been successfully applied to real-world applications over the past decades. Zhou et al. (2002) proposed an automatic pathological diagnosis procedure, which identifies lung cancer cells in the images based on NN ensembles. Jeong and Kim (2005) compared the prediction accuracy of single NN and NN ensemble for the rainfall–runoff model that predicts monthly inflows of dams. The NN ensemble used in this study uses the bootstrap method, which generates a training subset by sampling from the original training set with replacement and derives a final output by averaging individual predictions. This study reported that NN ensemble can further reduce generalization error compared with single NN. In the field of finance, Tsai and Wu (2008) used an NN ensemble that selects results with more than half votes for the problem of bankruptcy prediction and credit scoring. Zaier et al. (2010)

implemented six NN ensembles to model lake ice thickness and compared the estimation accuracy of ensembles with single NN models. The comparison results show that stacked generalization combining high- and low-level models significantly improves the predictive capability. Linares-Rodriguez et al. (2013) built an NN ensemble that averages five optimized NNs to estimate daily global solar radiation over large areas. Shao et al. (2014) modeled NN ensemble for fault diagnosis of proton exchange membrane fuel cell systems. In this study, sub-NN models are generated to minimize the correlation among the inputs of the sub-NNs, and weight coefficients are optimized to increase the accuracy of the ensemble. Alobaidi et al. (2014) proposed an improved NN ensemble using median statistic for spatial and temporal solar irradiance mapping. Berkhahn et al. (2019) used an NN ensemble through mean predictions of NNs generated with random initial weights to forecast real-time urban flooding. Khwaja et al. (2020) proposed an NN ensemble method that averages forecasted loads obtained from ensembles of NNs in parallel. In the proposed method, each ensemble consists of sequentially trained NNs based on bootstrapped samples. Moreira et al. (2021) employed a NN ensemble that considers the ensemble weights calculated through mixture analysis to forecast medium-term photovoltaic generation.

## 3. Proposed method

Various ensembles that combine prediction values have been developed to reduce the large prediction variance of NN. The prediction accuracy of NN becomes unstable for highly nonlinear problems with insufficient data set; therefore, determining the final prediction value by selecting high-accuracy prediction values is important. Thus, this section proposes the frequency distribution-based ensemble, which identifies the prediction values expected to be close to the true prediction value. In addition, the adaptive sampling strategy for the proposed ensemble is introduced to enhance the predictive performance of ensemble model efficiently by sequentially adding samples. The information flow of the proposed method is shown in Fig. 1. The frequency distribution-based ensemble comprises an ensemble preparation process and a core prediction value search algorithm. For a population data set, the adaptive sampling strategy uses the core prediction search algorithm to identify the next best sample point and update the original data set. The detailed explanations for the proposed method will be provided in the following sections.

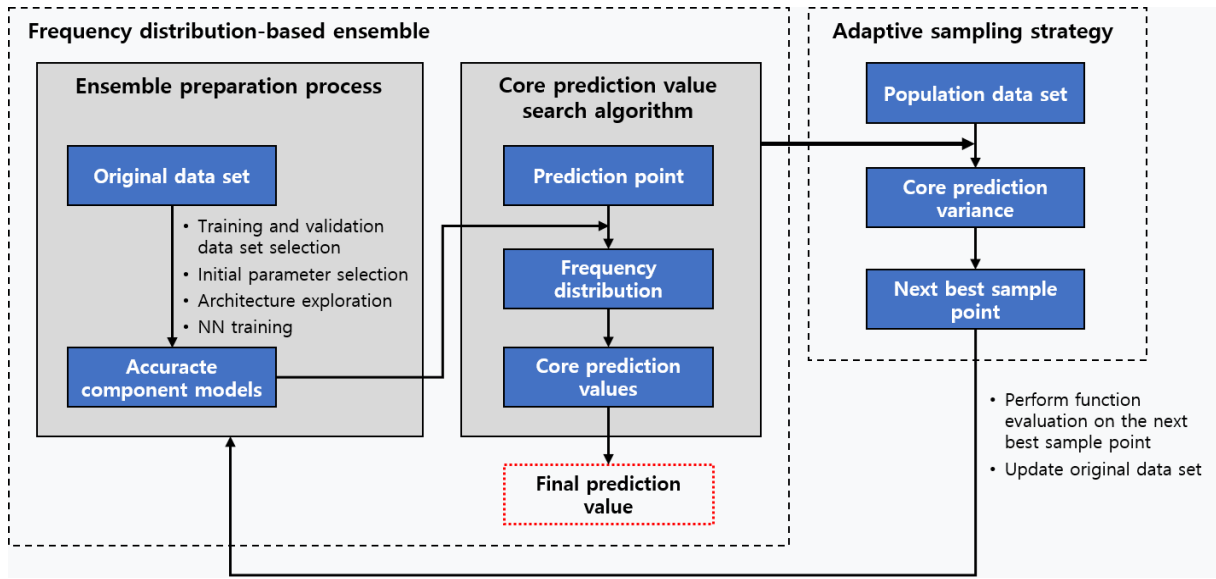

**Fig. 1** Information flow of the proposed method

### 3.1. Frequency distribution-based ensemble

For NN ensemble, an ensemble preparation process that generates diverse component models with high predictive performance is required. Once sufficiently diverse and accurate component models are available, the proposed frequency distribution-based ensemble identifies core prediction values, which are expected to be concentrated near the true prediction values. A core prediction value search algorithm is proposed to identify core prediction values. In this study, the target of NN ensemble is regression problems, and mean square error (MSE) is used as a performance metric to measure error in training and validation processes.

### 3.1.1 Ensemble preparation process

Component models that are sufficiently diverse and have high predictive performance are required to perform ensemble. In the ensemble preparation process, various component models are created in accordance with the combination of training and validation data set and the selection of initial parameters, and models with high predictive performance are selected among them. The flowchart of the ensemble preparation process is shown in Fig. 2. Searching for the architecture of NN is required before generating component models because the architecture of NN has a considerable influence on predictive accuracy (Elsken et al. 2019). An architecture that results in small training and validation errors for various combinations of training and validation data sets is selected. $m$ combinations of training and validation data sets are generated after determining the architecture of the NN to create $m$ component models. The combinations are generated on the basis of random sampling to prevent bias in training and validation data sets. For each combination, generating the NN model with random initial parameters is repeated 100 times to generate 100 NN models. The model with the smallest sum of training and validation errors among 100 NN models is then estimated to be accurate and saved as a component model for the combination. Finally, this process is repeated for *m* combinations of training and validation data sets, and *m* accurate component models are obtained. In this study, 100 component models considered to be sufficient are assumed for ensemble; thus, 100 combinations of training and validation data sets are used (Kourentzes et al. 2014).

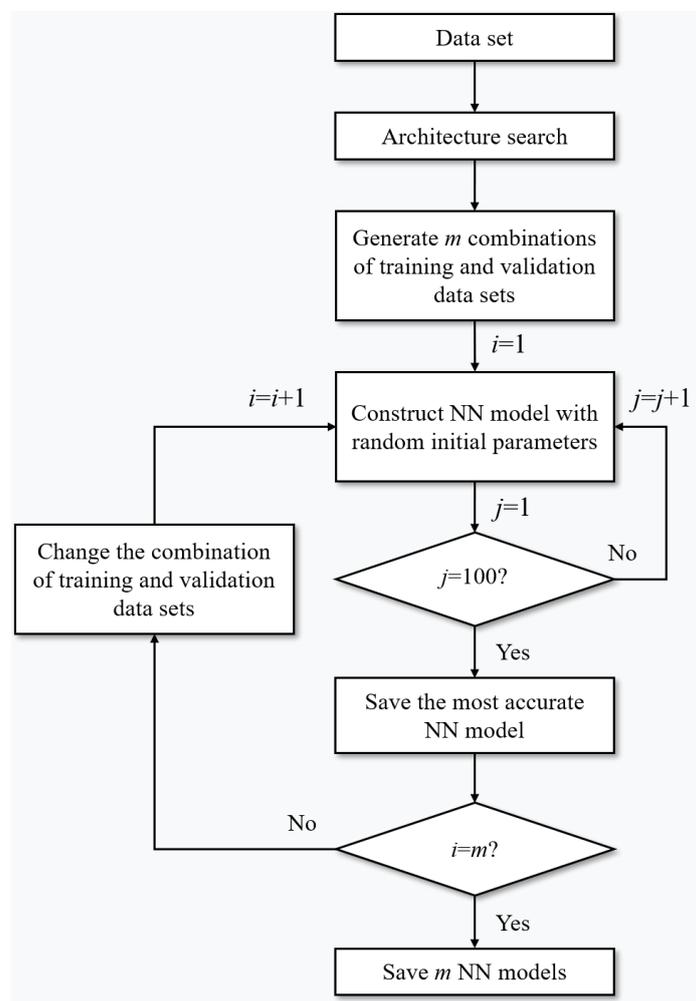

**Fig. 2** Flowchart of the ensemble preparation process

### 3.1.2 Core prediction value search algorithm

This section presents the proposed core prediction value search algorithm that identifies core prediction values, which are supported by multiple component models and expected to be concentrated near the true prediction values. Each component model can be regarded as a local model specialized to a specific training data set used for training. The prediction value supported by a small number of component models is unlikely to be close to the true prediction value and can be excluded from the ensemble. The basic idea of the frequency distribution-based ensemble is to include only highly correlated core prediction values that can be identified through statistical analysis of the prediction values. The frequency distribution based on the prediction value distribution shows different types of distributions depending on the bin width. Fig. 3 demonstrates an example of 100 prediction values derived from 100 component models for a given prediction point and the frequency distribution of 100 prediction values with 100 bins. The equation and prediction point used for the example are given as

$$g = 2x_1^2 - 1.05x_1^4 + \frac{x_1^6}{6} + x_1 x_2 + x_2^2, \ x_i = [1.3752, -2.2960]. \tag{1}$$

The number of samples in the data set used to build the NN model is 50, and the ensemble preparation process described in Section 3.1.1 is used to generate 100 component models. In the figure, a number of prediction values are located in the vicinity of the true prediction value. The correlation increases as the prediction values become close to each other, and prediction values with high correlation can be classified among all prediction values even if the true prediction value is unknown. In the frequency distribution, the most frequent bin identifies highly correlated prediction values and contains additional prediction values as the bin width increases. If the bin width is too small, then the prediction values that are expected to be close to the true prediction value cannot be sufficiently captured. This insufficiency can be solved by increasing the bin width. The performance of the frequency distribution-based ensemble becomes notable in the presence of outlier prediction values due to large nonlinearity and insufficient number of samples as in the example.

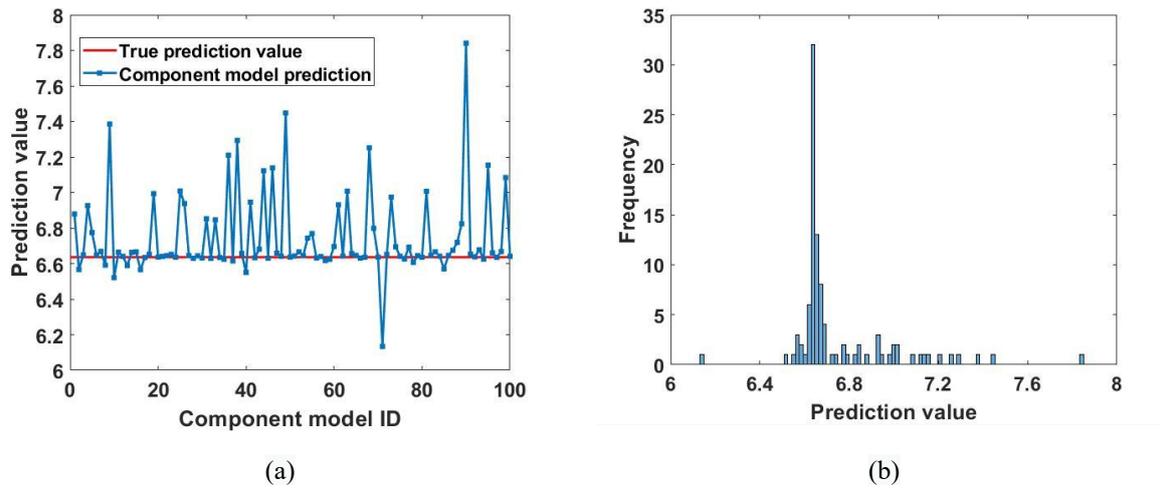

**Fig. 3** Example of 100 prediction values: (a) comparison between 100 prediction values and true prediction value, (b) frequency distribution of 100 prediction values with 100 bins

The flowchart of the frequency distribution-based ensemble is shown in Fig. 4. The following steps explain the frequency distribution-based ensemble employing the core prediction value search algorithm.

Step 1. Prepare $m$ prediction values using $m$ NN models generated from the ensemble preparation process.

Step 2. Set a minimum frequency criterion that the frequency of the most frequent bin should satisfy.

Step 3. Initialize the interval of frequency distribution using the number of prediction values as the number of bins.

Step 4. Generate frequency distribution with $m$ prediction values.

Step 5. If the frequency of the most frequent bin does not satisfy the minimum frequency criterion, then increase the bin width and return to Step 4. If the minimum frequency criterion is satisfied, then identify the range of the most frequent bin.

Step 6. Average of the prediction values within the range as the final prediction value.

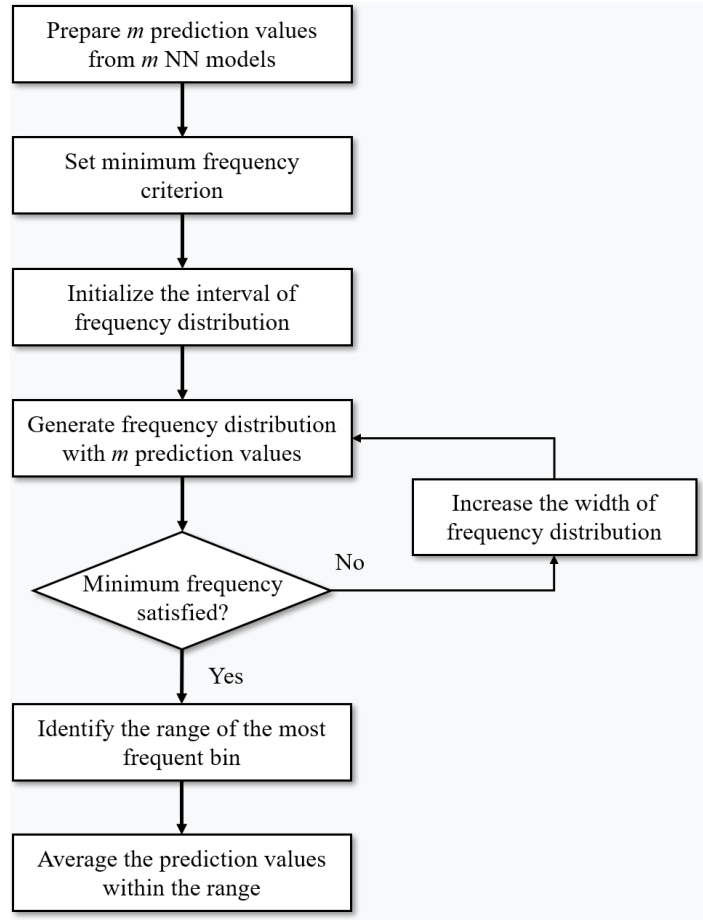

**Fig. 4** Flowchart of the frequency distribution-based ensemble

### 3.2. Adaptive sampling strategy for frequency distribution-based ensemble

In this section, a new adaptive sampling strategy, that is, the core prediction variance-based strategy, is proposed to improve the predictive performance of the proposed frequency distribution-based ensemble efficiently. In the field of Kriging, various learning functions and update schemes have been developed to reduce the uncertainty of prediction by adaptively adding samples based on Kriging variance (Echard et al. 2011). An individual NN model cannot quantify the prediction uncertainty; however, ensembles can derive the prediction variance for each prediction point through the prediction values obtained from component models (Yu et al. 2008). The prediction variance is used to determine the next best sample point, and adding samples to the sample points with large prediction variance can reduce the overall prediction variance of the ensemble model (Eason and Cremaschi 2014).

The frequency distribution-based ensemble makes predictions based on core prediction values. Thus, a new core prediction variance-based strategy that adds samples based on a core prediction variance indicating the variance of the core prediction values can be proposed. Prediction values supported by only a few number of component models can be considered outliers and must thus be excluded in quantifying the uncertainty of the prediction point. Therefore, the core prediction variance-based strategy quantifies the uncertainty of the prediction point based on the core prediction values and explores prediction points with high uncertainty despite the support of multiple component models. The proposed strategy can effectively increase the prediction accuracy of the ensemble model by adding samples to the prediction points with high prediction uncertainty. The core prediction variance-based strategy, which sequentially adds samples based on the core prediction variance, comprises the following procedures.

    Step 1. Generate population $S$ throughout the design space.

    Step 2. Calculate the core prediction variance for all the sample points in $S$ by using the ensemble model.

    Step 3. Identify the sample point with the highest core prediction variance as the next best sample point.

    Step 4. Perform function evaluation on the next best sample point and update the original data set.

    Step 5. Construct the ensemble model according to the updated sample set.

## 4. Case studies

The effectiveness of the proposed frequency distribution-based ensemble and adaptive sampling strategy is verified in this section through various case studies. For comparison, explanations of existing ensemble methods and adaptive sampling strategies are provided.

### 4.1. Surrogate model accuracy

This section presents the results obtained from various surrogate modeling methods, including Kriging and ensembles. The existing ensemble methods compared in this section are average, weighted, and mode ensembles and are described in Section 4.1.1. A total of 500 test sets are used and five surrogate modeling methods are compared to investigate the prediction accuracy of surrogate modeling methods: (1) Method 1 is Kriging; (2) Method 2 is average ensemble; (3) Method 3 is weighted ensemble; (4) Method 4 is mode ensemble; and (5) Method 5 is the proposed frequency distribution-based ensemble. For an accuracy metric of each method, normalized root mean square error (NRMSE) is calculated as

$$\mathrm{NRMSE} = \frac{1}{\bar{y}} \sqrt{\frac{\sum_{k=1}^{n_{test}} (y_k - \hat{y}_k)^2}{n_{test}}}, \tag{2}$$

where $n_{test}$ is the number of test sets, $y_k$ is the true prediction value of the $k$th test set, $\hat{y}_k$ is the estimated value of the $k$th test set, and $\bar{y}$ is the mean of true prediction values. In ensemble methods, the ratios for training and validation sets are 0.8 and 0.2, respectively. The hyperbolic tangent function is used as the activation function, and the Levenberg–Marquardt (LM) algorithm is used for optimization. The NN training is conducted using a standard desktop (Intel XEON CPU @ 192.0 GB GPU and 2.70 GHz).

**4.1.1 Existing ensemble methods**

**4.1.1.1 Average ensemble**

The average ensemble is the simplest way to combine prediction values and derives a final prediction by averaging predictions of component models (James et al. 2013). The average ensemble can be defined as

$$\tilde{y}^{Average} = \frac{1}{m}\sum_{i=1}^{m} y_i, \tag{3}$$

where $\tilde{y}^{Average}$ is the average ensemble value, and $y_i$ indicates the prediction value of the $i$th component model. The average ensemble reduces the risk of overfitting and improves the predictive performance compared with a single model but demonstrates sensitivity to outliers. In addition, the predictive performance can deteriorate for skewed distributions.

**4.1.1.2 Weighted ensemble**

The weighted ensemble is a method to improve the predictive performance by providing some component models with larger weights than others (Bishop 1995). The weights are determined in accordance with the predictive accuracy because accurate component models are more reliable than inaccurate ones (Opitz and Shavlik 1996). The weighted ensemble can be defined as

$$\tilde{y}^{Weighted} = \sum_{i=1}^{m} w_i y_i, \quad \sum_{i=1}^{m} w_i = 1, \tag{4}$$

where $\tilde{y}^{Weighted}$ is the weighted ensemble value, and $w_i$ indicates the weight of the $i$th component model. The component model with a small validation error is generally considered to have high predictive performance and is given a high weight (Zhang et al. 2022). However, high weight is given to a component model with a small sum of training and validation errors in this study to prevent a poorly trained component model with a low validation error from receiving high weight. Finally, the weights can be calculated as (Goel et al. 2007)

$$w_i = \frac{w_i^*}{\sum_{j=1}^{m} w_j^*}, \ w_i^* = (e_i + 0.05 e_{avg})^{-1}, \ e_{avg} = \frac{1}{m}\sum_{j=1}^{m} e_j, \tag{5}$$

where $e_i$ is the sum of training and validation errors of the $i$th component model.

**4.1.1.3 Mode ensemble**

The mode defined as the most frequent value in a data set can be insensitive to outliers. Kourentzes et al. (2014) proposed the mode ensemble that derives a mode using KDE to perform ensemble. KDE is a nonparametric technique for estimating an unknown probability density function, in which previously known density functions of the observed data are averaged. For m prediction values, the kernel density estimator can be defined as

$$w_i = \frac{w_i^*}{\sum_{j=1}^{m} w_j^*}, \ w_i^* = (e_i + 0.05 e_{avg})^{-1}, \ e_{avg} = \frac{1}{m}\sum_{j=1}^{m} e_j, \tag{6}$$

where $K$ is the kernel function, and $h > 0$ is the bandwidth adjusting the smoothness of the distribution. The Gaussian kernel, which is widely used for computational reasons, is defined as

$$\phi_h(x) = \frac{1}{h\sqrt{2\pi}} e^{-\frac{x^2}{2h^2}}. \tag{7}$$

The smoothing effect increases with the bandwidth, and providing an appropriate bandwidth for the data set is required. The optimal bandwidth that minimizes the mean integrated squared error is given by (Silverman 2018)

$$\phi_h(x) = \frac{1}{h\sqrt{2\pi}} e^{-\frac{x^2}{2h^2}}, \tag{8}$$

where sigma indicates the standard deviation of $m$ prediction values used for the ensemble. The value of $x$ corresponding to the maximum density is the mode of the prediction value distribution and is used as the mode ensemble value $y$. If the prediction value distribution is multimodal, then a mode close to the prediction values can be selected as the mode ensemble value.

**4.1.2 Numerical example I: 2D mathematical problem**

A 2D mathematical problem is formulated as shown below to compare the results of a highly nonlinear problem with insufficient number of samples.

$$g = 2x_1^2 - 1.05 x_1^4 + \frac{x_1^6}{6} + x_1 x_2 + x_2^2, \ x_i \in [-3, 3]. \tag{9}$$

The number of samples in the data set used to build the surrogate model is 50, and the random variables $x_i$ are uniform random variables. NRMSE results obtained using various minimum frequency criteria are compared to determine the minimum frequency criterion of Method 5, and the results are shown in Table 1. The results confirmed that the prediction accuracy is highest when the frequency of the most frequent bin is larger than or equal to 20% of the total bins.

Table 1 The NRMSE results obtained using various minimum frequency criteria (Method 5)

| 10% | 20% | 30% | 40% | 50% | 60% | 70% | 80% | 90% | 100% |
|---|---|---|---|---|---|---|---|---|---|
| 0.0033 | 0.0032 | 0.0040 | 0.0064 | 0.0062 | 0.0087 | 0.0100 | 0.0134 | 0.0163 | 0.0167 |

The NRMSE results obtained from the five methods are shown in Table 2, and the box plots of relative errors are plotted in Fig. 5. In the box plots, the median values of Methods 1, 2, 3, 4, and 5 are 1.7218, 1.4514, 0.5879, 0.1824, and 0.0682, respectively. The results verify that the prediction accuracy of the proposed frequency distribution-based ensemble is the most accurate. The prediction accuracy of Method 1 (Kriging) is lower than that of ensemble methods because the interpolation-based Kriging model becomes significantly inaccurate when the problem is highly nonlinear and the number of samples is insufficient. Among ensemble methods, the prediction accuracy of the average ensemble is low. This finding is due to the inclusion of all biased outliers of average ensemble in the prediction values. The weighted ensemble can cope with outliers better than the average ensemble because different weights are given to the prediction values according to the predictive performance of the component model. However, the prediction accuracy is lower than Methods 4 and 5 because outliers are still considered in the ensemble. Thus, Methods 4 and 5 can exclude outliers in the ensemble because these methods focus on frequent prediction values. Compared with Method 4, Method 5 further improves the prediction accuracy. Method 4 considers only the mode, whereas Method 5 considers the prediction values around the mode as well as the mode itself, thereby coping with the uncertainty of the mode.

Table 2 The NRMSE results of 2D mathematical problem

| Method 1 | Method 2 | Method 3 | Method 4 | Method 5 |
|---|---|---|---|---|
| 0.0658 | 0.0168 | 0.0109 | 0.0067 | 0.0032 |

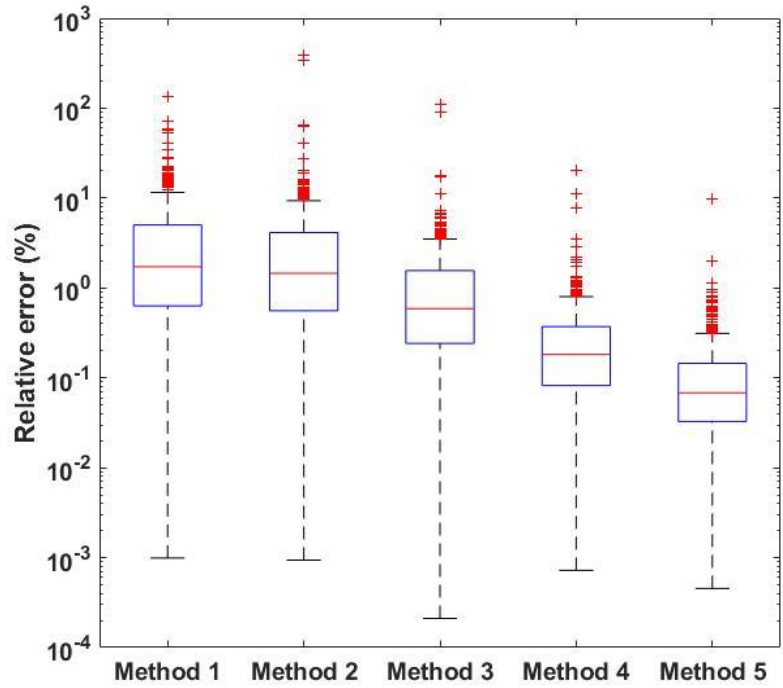

**Fig. 5** Box plots of relative errors obtained from 2D mathematical problem

The NRMSE results obtained using various numbers of samples in the data set are compared in Table 3 to further investigate the impact of the number of samples in the data set used to build the surrogate model. If the number of samples is insufficient, then the prediction accuracy of each component model decreases and overfitting occurs, resulting in large errors in ensemble methods. In the case of Method 5, the prediction accuracy is lower than other ensemble methods because the ensemble is performed by collecting the prediction values with large errors. However, results confirmed that the prediction accuracy of Method 5 is significantly improved as the number of samples and the prediction accuracy of each component model increase.

**Table 3** The NRMSE results obtained using various number of samples in the data set

|  | Number of samples in the data set | | | | |
| --- | --- | --- | --- | --- | --- |
|  | 10 | 20 | 30 | 40 | 50 |
| Method 1 | 1.0759 | 0.6426 | 0.2795 | 0.1806 | 0.0658 |
| Method 2 | 1.1911 | 0.4069 | 0.1570 | 0.0345 | 0.0168 |
| Method 3 | 1.1805 | 0.3353 | 0.1533 | 0.0266 | 0.0109 |
| Method 4 | 1.1698 | 0.4478 | 0.1593 | 0.0156 | 0.0067 |
| Method 5 | 1.2148 | 0.4811 | 0.1655 | 0.0162 | 0.0032 |

### 4.1.3 Numerical example II: Modified 4D Powell function

A modified 4D Powell function with severe nonlinearity is introduced in this section to compare the

prediction accuracy of the five methods. The modified Powell function is formulated as

$$g = (x_{4i-3} + 10x_{4i-2})^2 + 5(x_{4i-1} - x_{4i})^2 + (x_{4i-2} - 2x_{4i-1})^6 + 10(x_{4i-3} - x_{4i})^4, \ x_i \in [0,10]. \tag{10}$$

The number of samples in the data set used to build the surrogate model is 70, and the random variables $x_i$ are uniform random variables. The NRMSE results of the modified 4D Powell function are shown in Table 4, and the box plots of relative errors are plotted in Fig. 6. In the box plots, the median values of Methods 1, 2, 3, 4, and 5 are 455.8085, 7.4849, 6.7684, 4.7140, and 3.6897, respectively. The number of samples compared with the dimension of the problem is insufficient and the nonlinearity of the problem is high; thus, the prediction accuracy of Method 1 (Kriging) becomes significantly lower than that of ensemble methods. For a given prediction point, ensemble methods derive the prediction value with high accuracy by using various component models with high predictive performance. Methods 2 (average ensemble) and 3 (weighted ensemble) show similar accuracy. In Method 3, weights that depend only on the global accuracy of the component model are given to the prediction values obtained from a given prediction point; therefore, the prediction accuracy of component models according to the prediction point cannot be reflected. Methods 4 and 5, which select different component models according to the prediction point, show high prediction accuracy, and the proposed Method 5 shows the highest prediction accuracy.

Table 4 The NRMSE results of 4D Powell function

| Method 1 | Method 2 | Method 3 | Method 4 | Method 5 |
| --- | --- | --- | --- | --- |
| 0.9028 | 0.1195 | 0.1209 | 0.0516 | 0.0466 |

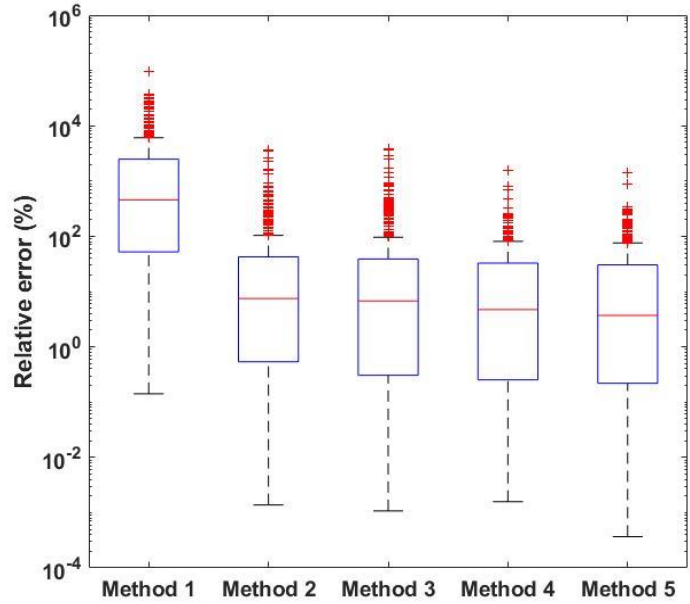

**Fig. 6** Box plots of relative errors obtained from 4D Powell function

### 4.1.4 Numerical example III: 8D Borehole model

The borehole model used to describe the flow of water through a borehole is given as (Joseph et al. 2008)

$$g = \frac{2\pi T_u (H_u - H_l)}{\ln(r/r_w)\left[1 + \frac{2LT_u}{\ln(r/r_w)r_w^2 K_w} + \frac{T_u}{T_l}\right]}, \tag{11}$$

where the response is the flow rate of water, and the ranges of the input variables are presented in Table 5. The number of samples in the data set used to build the surrogate model is 120, and all the input variables are extracted from uniform distributions. The NRMSE results of the 8D borehole model are shown in Table 6, and the box plots of relative errors are plotted in Fig. 7. In the box plots, the median values of Methods 1, 2, 3, 4, and 5 are 10.8631, 0.5950, 0.4174, 0.3203, and 0.3123, respectively. The NRMSE results verify that Method 5 has higher prediction accuracy than other methods.

**Table 5** The ranges of the input variables

| Input variables | Description | Range |
| --- | --- | --- |
| $r_w$ | Radius of borehole ($m$) | [0.05, 2] |
| $R$ | Radius of influence ($m$) | [100, 50000] |
| $T_u$ | Transmissivity of upper aquifer ($m^2/yr$) | [63070, 115600] |
| $H_u$ | Potentiometric head of upper aquifer ($m$) | [990, 1110] |
| $T_l$ | Transmissivity of lower aquifer ($m^2/yr$) | [63.1, 116] |
| $H_l$ | Potentiometric head of lower aquifer ($m$) | [700, 820] |
| *L* | Length of borehole (*m*) | [1120, 2680] |
| $K_w$ | Hydraulic conductivity of borehole (*m/yr*) | [1500, 30000] |

**Table 6** The NRMSE results of 8D borehole model

| Method 1 | Method 2 | Method 3 | Method 4 | Method 5 |
|---|---|---|---|---|
| 0.2103 | 0.0172 | 0.0127 | 0.0099 | 0.0083 |

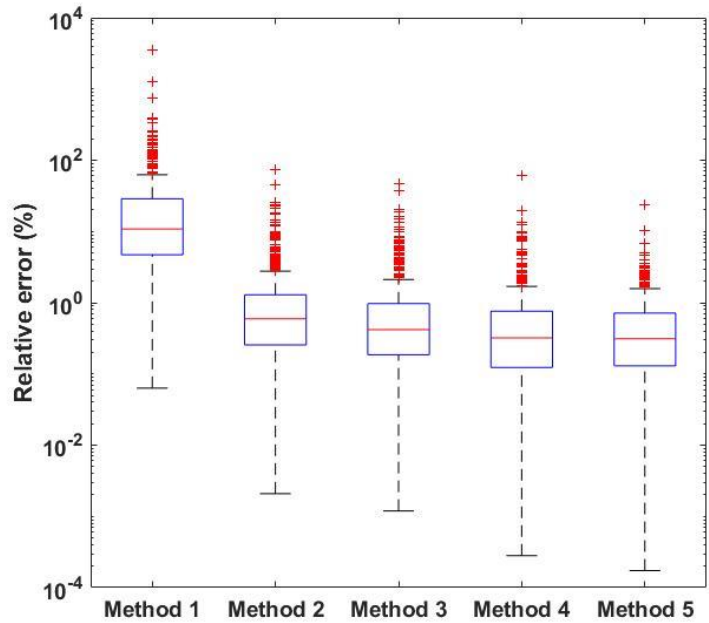

**Fig. 7** Box plots of relative errors obtained from 8D borehole model

**4.1.5 Engineering example: 6D arm model**

This section presents an engineering example predicting the maximum von Mises stress that occurs in a 6D arm model (Altair 2017). The loading and boundary conditions of the model are presented in Fig. 8, and the ranges of input variables are shown in Table 7. The morphing parameters indicated by "*dv*" constitute the input variables and control the geometry of the arm. The number of samples in the data set used to build the surrogate model is 100, and all the input variables are extracted from uniform distributions. The NRMSE results of the 6D arm model are shown in Table 8, and the box plots of relative errors are plotted in Fig. 9. In the box plots, the median values of Methods 1, 2, 3, 4, and 5 are 6.4123, 0.8602, 0.8627, 0.8362, and 0.8273, respectively. The nonlinearity of this example is moderate; thus, the difference in prediction accuracy between ensemble methods is small. The NRMSE results show that Method 5 has higher prediction accuracy than other methods, verifying that the proposed Method 5 can improve the prediction accuracy compared with other ensemble methods even in engineering examples.

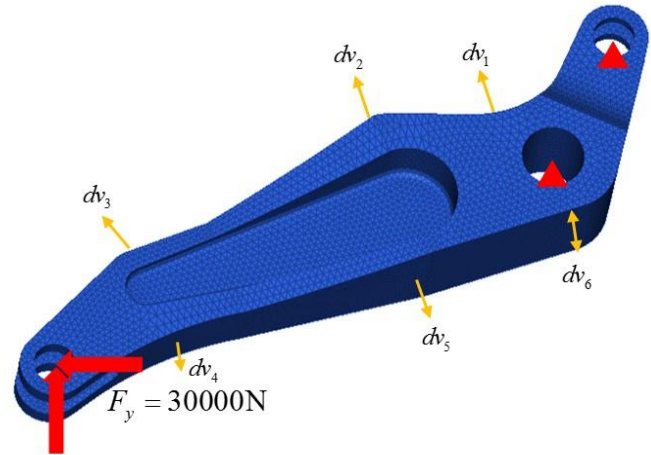

**Fig. 8** Loading and boundary conditions of 6D arm model

**Table 7** Properties of input variables of 6D arm model

| Input variables | Range |
|---|---|
| $dv_1$ | [−0.5, 4] |
| $dv_2$ | [0, 3] |
| $dv_3$ | [−1, 5] |
| $dv_4$ | [−1, 1] |
| $dv_5$ | [−1, 5] |
| $dv_6$ | [−1, 4] |

**Table 8** The NRMSE results of 6D arm model

| Method 1 | Method 2 | Method 3 | Method 4 | Method 5 |
|---|---|---|---|---|
| 0.1249 | 0.0207 | 0.0206 | 0.0189 | 0.0182 |

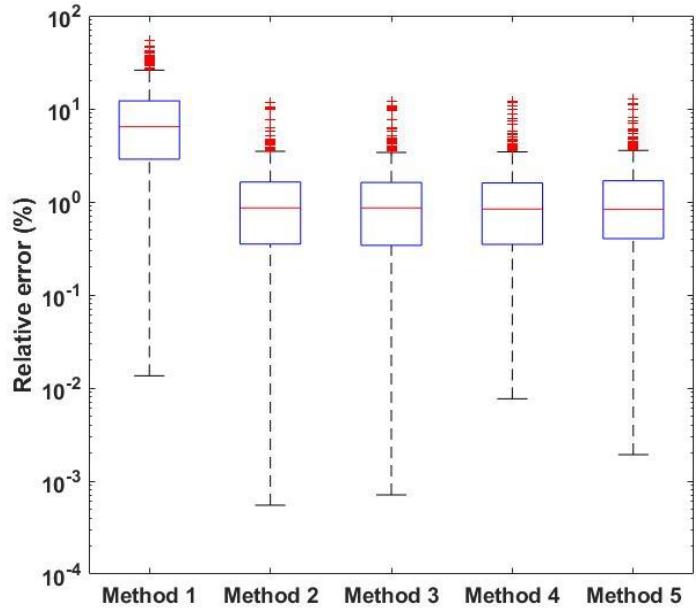

**Fig. 9** Box plots of relative errors obtained from the 6D arm model

**4.2. Adaptive sampling strategy for frequency distribution-based ensemble**

In this section, the proposed core prediction variance-based strategy presented in Section 3.2 is verified using several examples. For comparison, the previously developed space-filling strategy and the prediction variance-based strategy are compared. These existing adaptive sampling strategies are described in Section 4.2.1. One sample is sequentially added in accordance with each adaptive sampling strategy to verify the performance of each adaptive sampling strategy, and the prediction accuracy of the updated frequency distribution-based ensemble is compared through NRMSE.

**4.2.1 Existing adaptive sampling strategies**

**4.2.1.1 Incremental Latin hypercube sampling ($i$LHS)**

Latin hypercube sampling (LHS) is a statistical sampling method that performs near-random sampling from multidimensional distributions (McKay et al. 1979). In LHS, the range of each variable is divided into $n$ equivalent intervals, where $n$ is the number of sample points to be generated (Eason and Cremaschi 2014). $n$ coordinates are obtained for each variable by selecting one coordinate in each interval. One sample point is then constructed by randomly extracting one coordinate from each variable, and distributed $n$ sample points can be obtained by repeating the procedure $n$ times. Thus, the probability that a sample point belongs to each interval is equal for all variables. If a new sample point is added, then some intervals will be overfilled and the equal probability condition for intervals will be violated. Therefore, a new run of the LHS for a new number of samples is required. The *i*LHS is a space-filling strategy that enhances the LHS to satisfy the equal probability condition of the LHS even for an increased number of samples (Nuchitprasittichai and Cremaschi 2013). The *i*LHS divides the range of the variable into equivalent intervals with a new number of samples, discards overlapping samples, and fills in new sample points to maintain the equal probability condition of LHS. Samples can be added sequentially through the *i*LHS to improve the prediction accuracy of ensembles.

**4.2.1.2 Prediction variance-based strategy**

Ensembles make predictions using various component models; thus, different prediction values are derived for a given prediction point, and the prediction variance occurs. The prediction variance represents the prediction uncertainty of the prediction point and is used to determine the next best sample point (Eason and Cremaschi 2014). Adding samples to sample points with large prediction variance can reduce the overall prediction variance of the ensemble model. The prediction variance-based strategy that sequentially adds samples based on the prediction variance has procedures similar to those presented in Section 3.2. In Step 2, the prediction variance is calculated for all the sample points; in Step 3, the sample point with the highest prediction variance is identified as the next best sample point. The prediction variance-based strategy differs from the proposed core prediction variance-based strategy in that it determines the next best sample point based on prediction values derived from all component models. Therefore, the prediction variance may contains errors due to component models with low accuracy, providing an inappropriate next best sample point.

**4.3. Case studies on adaptive sampling strategy**

Adaptive sampling strategies that improve the predictive performance of the frequency distribution-based ensemble by sequentially adding samples are compared in this section through case studies. The predictive performance of the frequency distribution-based ensemble, which is improved as samples are added, is the target of comparison. The *i*LHS, prediction variance-based strategy, and the proposed core prediction variance-based strategy are compared.

**4.3.1 Numerical example: 2D mathematical problem**

Three adaptive sampling strategies are applied for the 2D mathematical problem presented in Section 4.1.2.

Initially, the number of samples in the data set used to build the frequency distribution-based ensemble is 50, and samples are sequentially added in accordance with the adaptive sampling strategy until the total number of samples reaches 100. The NRMSE results according to the number of samples are presented in Fig. 10. In all three strategies, the predictive performance of the frequency distribution-based ensemble increases as samples are added. The results show that the proposed core prediction variance-based strategy most efficiently increases the predictive performance of the frequency distribution-based ensemble. In the case of $i$LHS, the improvement of prediction accuracy is relatively low because the prediction uncertainty at each prediction point is disregarded and samples are added randomly. The results verify that adding samples considering the core prediction variance rather than the prediction variance improves the predictive performance of the frequency distribution-based ensemble efficiently. In the early stage of adding samples, the prediction variance- and core prediction variance-based strategies show similar prediction accuracy; however, the difference in the prediction accuracy between the two becomes clear as additional samples are included. When the number of samples is 100, the core prediction variance-based strategy shows 74.7% and 59.4% lower NRMSE than $i$LHS and prediction variance-based strategy, respectively. In addition, to reach the NRMSE of the core prediction variance-based strategy obtained with 100 samples, $i$LHS requires 287 samples and the prediction variance-based strategy requires 219 samples.

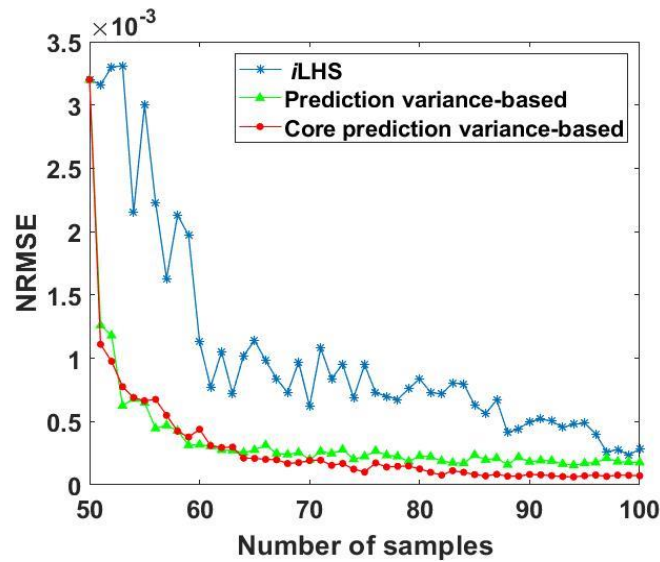

Fig. 10 The NRMSE results according to the number of samples (2D mathematical problem)

### 4.3.2 Engineering example: 6D arm model

In this section, three adaptive sampling strategies are compared for the 6D arm model presented in Section

4.1.5. Initially, 100 samples are used to build the frequency distribution-based ensemble, and 50 samples are added sequentially according to the adaptive sampling strategy. The NRMSE results according to the number of samples are presented in Fig. 11. The results verify that the proposed core prediction variance-based strategy most efficiently improves the predictive performance of the frequency distribution-based ensemble in the engineering example. As in Section 4.3.1, the RMSE difference between the core prediction variance- and prediction variance-based strategies becomes clear as samples are added. When the number of samples is 100, the core prediction variance-based strategy shows 31.6% and 12.3% lower NRMSE than $i$LHS and prediction variance-based strategy, respectively. To reach the NRMSE of the core prediction variance-based strategy obtained with 150 samples, $i$LHS requires 196 samples and the prediction variance-based strategy requires 189 samples. In the case of engineering applications, high computational cost is involved in performing simulation. Therefore, the proposed core prediction variance-based strategy helps obtain a frequency distribution-based ensemble model with high predictive performance using relatively low computational effort.

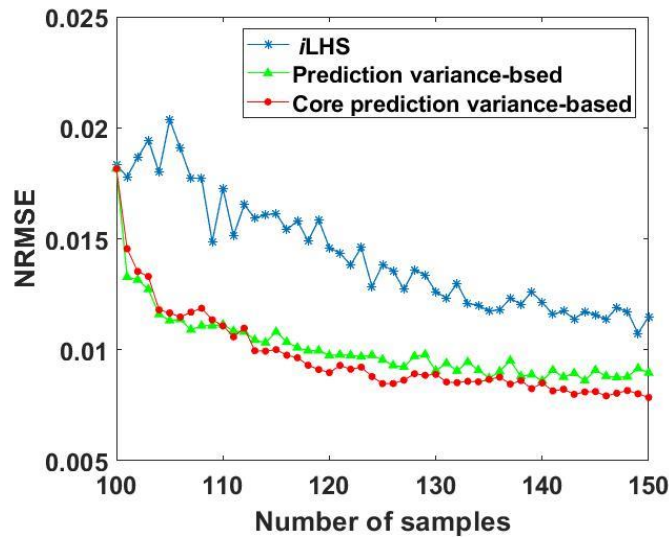

**Fig. 11** The NRMSE results according to the number of samples (6D arm model)

## 5. Conclusion

NNs are considered promising surrogate models with good prediction accuracy, and various ensembles have been developed to reduce the large prediction variance of NN. This study proposes the frequency distribution-based ensemble that identifies core prediction values, which are expected to be concentrated near the true prediction value, and derives the average of the core prediction values as the final prediction value, to improve the predictive performance of ensembles for highly nonlinear problems with insufficient data set. In the ensemble preparation process, various component models are generated in accordance with the combination of training and validation data sets and the selection of initial parameters, and models with high predictive performance are selected. For a given prediction point, various prediction values can be obtained from the component models, and the frequency distribution-based ensemble performs statistical analysis with the frequency distribution generated on the basis of the prediction values to classify the concentrated prediction values. A core prediction variance-based strategy, which sequentially adds samples based on the core prediction variance, is proposed to improve the predictive performance of the proposed frequency distribution-based ensemble efficiently. The core prediction variance is the variance of the core prediction values. After all the core prediction variances are calculated for the generated population, the sample point with the highest core prediction variance becomes the next best sample point. Various case studies verified that the prediction accuracy of the frequency distribution-based ensemble is higher than that of Kriging and other existing ensemble methods and the core prediction variance-based strategy efficiently improves the predictive performance of the frequency distribution-based ensemble.

The contribution of this study can be summarized as follows. First, the proposed frequency distribution-based ensemble can improve the predictive performance by excluding outlier prediction values for highly nonlinear problems with insufficient data set. Second, the proposed frequency distribution-based ensemble can cope with the uncertainty of the most frequent value by adaptively identifying the most frequent bin. Third, the proposed core prediction variance-based strategy efficiently improves the predictive performance of the frequency distribution-based ensemble by calculating the core prediction variance and sequentially adding samples to the point where the actual prediction error is expected to be large.

This study also has some limitations. The nonlinearities and properties of the problems required for the frequency distribution-based ensemble to work efficiently should be investigated quantitatively. In addition, the effect of the minimum frequency criterion on the prediction accuracy should be studied. In the future work, the frequency distribution-based ensemble can be applied to increase the efficiency and accuracy of reliability analysis. A learning function that recommends the sample point near the limit state with a large core prediction variance as the next best sample point must be developed to increase the accuracy of frequency distribution-based ensemble efficiently.

**Acknowledgments**

This work was supported by the National Research Foundation of Korea (2018R1A5A7025409) and the Ministry of Science and ICT of Korea (No.2022-0-00969).